\documentclass[12pt]{article}
\usepackage{amsmath}
\usepackage{authblk}
\usepackage[american]{babel}
\usepackage{booktabs}
\usepackage{colortbl}
\usepackage{courier}
\usepackage{csquotes}
\usepackage{float}
\usepackage{geometry}
\usepackage{graphicx}
\usepackage{indentfirst}
\usepackage{longtable}
\usepackage{multirow}
\usepackage{pdflscape}
\usepackage{hyperref}
\usepackage[style=apa, sorting=nyt]{biblatex}
\DeclareLanguageMapping{american}{american-apa}

\newcommand{\courier}[1]{{\fontfamily{pcr}\selectfont #1}}

\title{\large \textbf{Empirical Prompt Engineering for Construct Identification with Large Language Models}}

\author[1]{Kylie L. Anglin}

\author[2]{Stephanie Milan}

\author[1]{Brittney Hernandez}

\author[1]{Claudia Ventura}

\affil[1] {Department of Educational Psychology, Neag School of Education, University of Connecticut}
\affil[2] {Department of Psychological Sciences, College of Liberal Arts and Sciences, University of Connecticut}

\begin{document}
\maketitle

\begin{abstract}
Due to their architecture and vast pre-training data, large language models (LLMs) demonstrate strong performance on text classification tasks. However, LLM classifications are highly responsive to prompt wording, particularly, as we show, in domains like psychology, where constructs are often latent, complex, and theory driven. Here, we present and evaluate a systematic framework for improving psychological construct identification through prompt engineering. We combinatorially generate prompts by appending random selections of multiple variants of construct definitions, task instructions, coding guidance, and examples. Empirically selecting the highest performing of these combinations in a training dataset substantially improves alignment between LLM and human classifications. In contrast, prompting techniques such as personas, chain-of-thought reasoning, and explanations provide smaller and less consistent improvements. This finding holds across multiple models and constructs. Overall, the approach we describe offers a practical, systematic, and theory-aware method for increasing the alignment between human and LLM classifications in settings where validity is critical.
\end{abstract}

\section*{Author Note}

Kylie Anglin is the corresponding author and can be reached at: Email: Kylie.Anglin@uconn.edu; Phone: 860-486-0181. We have no known conflicts of interest to disclose. This research was supported by the Clinical Research and Innovation Seed Program (CRISP) at the University of Connecticut.

\pagebreak

\section*{Introduction}

Psychology is largely the study of internal processes—of thoughts, feelings, and mental states that occur within the mind. As such, psychological constructs of interest are often latent and cannot be directly observed. In clinical and research settings, however, these constructs may be inferred from what a client says, in writing or in speech (\cite{stavropoulos_shadows_2024}). For example, the client may reveal \textit{negative core beliefs} (overgeneralized judgments about themselves, others, or the world; \cite{beck_cognitive-behavioral_2011}; \cite{wenzel2012}) during a therapy session. Alternatively, they may demonstrate \textit{positive meaning making} (describing positive change or growth due to a negative experience; \cite{boals_use_2012}) when writing about a difficult experience. Thus, measuring these and other latent constructs often involves the labeling of text data. 

Commonly it is experts who generate this labeled data by combing through and classifying text segments. This process of hand-coding comes with substantial challenges, particularly as the size of the data scales, including: costs, delays, coder fatigue, and coder drift (\cite{can_it_2016}). Alternatively, researchers may train a supervised learning algorithm to classify unlabeled texts automatically based on patterns learned in the hand-labeled dataset (\cite{dehghani_tacit_2017}; \cite{semeraro_emoatlas_2023}; \cite{shin_investigating_2025}). This approach is eminently scalable – the cost of classifying an additional text after model training is negligible – but its effectiveness depends heavily on the quantity of labeled training data, with quantities in the thousands commonly required. 

Large language models (LLMs) bring a new option for researchers (e.g., \cite{martinez_scoring_2024}; \cite{van_genugten_automated_2024}). Given LLMs have been pre-trained on enormous corpora of language, they come to a new classification task with a substantial leg up compared to their traditional supervised counterparts. Through their pre-training, they have already learned linguistic patterns that capture diverse word meanings, grammatical structures, and idiomatic patterns (\cite{devlin_bert_2019}). This dramatically lowers the necessary number of labeled examples required to reach a high level of model accuracy when classifying new constructs. Indeed, using \textit{zero-shot classification}, a researcher can prompt an LLM to categorize a text without any context-specific labeled training data at all (\cite{reynolds_prompt_2021}), or, in \textit{few-shot classification} with only a few example classifications (\cite{brown_language_2020}).

Yet, there is no guarantee that either zero- or few-shot classifications will align with researchers’ understanding of the construct of interest. Thus, the validity of inferences resulting from these classifications (e.g., concluding that a text indicates a client has engaged in meaning making) must be established based on evidence. A core piece of evidence in favor of this validity is a high degree of alignment between LLM-based classifications and human classifications (\cite{bunt_validating_2025}). \textit{Prompt engineering}, the process of optimizing prompts to improve LLM performance on specific tasks, has emerged as an impactful lever for increasing this alignment (\cite{kojima_large_2022}; \cite{weber_evaluation_2023}; \cite{zhou2022}).

Unfortunately, the prompt engineering literature remains limited for psychological and behavioral researchers. First, the majority of the prompt engineering literature focuses on text generation (e.g., asking a model to answer an open-ended question) rather than text classification (where the model should select between a limited number of options in its response), leaving prompt engineering methods for text classification underdeveloped (\cite{white_prompt_2023}). Second, many commonly recommended prompt engineering approaches, like chain-of-thought prompting (\cite{wei_chain--thought_2022}), have less evidence of efficacy within the classification context (\cite{lampinen_can_2022}). Third, there is very little research testing the efficacy of prompt engineering techniques in psychological and behavioral sciences. Thus, it is often unclear how prompt engineering approaches will perform for the identification of complex, field-specific constructs (\cite{stavropoulos_shadows_2024}) where expert understandings of terms may differ from the largely non-academic documents on which the LLMs were trained (\cite{ziems_can_2024}). 

This article fills that gap by conducting a series of prompt engineering case studies, systematically testing over a thousand varied prompts for identifying psychological constructs across variable contexts. In doing so, we demonstrate how these methods and techniques may be rigorously applied and use the results to draw out recommendations for future researchers. Our case studies focus on three text classification tasks, one which we view as a low-inference task that closely matches a common general-purpose use of text classifiers (emotion identification), and two higher-inference tasks with precise, theory-driven definitions: identifying positive meaning making and negative core beliefs in participant writing samples. 

In our experiments, we distinguish between what we term baseline prompts and additive techniques. Baseline prompts simply describe the construct and classification task, while additive techniques are modifications that can be layered onto any baseline prompt regardless of the construct being classified. In total, we evaluate two approaches to generating baseline prompts (combinatorial empirical prompting and automatic prompt engineering), three additive prompting techniques (persona patterns, chain-of-thought, and explanations), and two uses of labeled examples (zero-shot and few-shot) across three constructs (gratitude, positive meaning making, and negative core beliefs). The breadth and scale of our experiments allow us to 1) assess state-of-the-art prompt engineering methods; 2) demonstrate a comprehensive LLM text classification process that prioritizes validity and validation; 3) test the robustness of these recommendations to variation in tasks and data; and 4) provide recommendations to the field. 

To empirically select an optimal baseline prompt, we break a codebook into key components (context, definition, task, and inclusion/exclusion criteria) and write multiple versions of each. Then, we generate many prompt variants by randomly combining these components (e.g., randomly selecting one context description, one construct definition, one task description, and a subset of inclusion/exclusion criteria). We evaluate each variant in a training dataset and select the highest-performing prompt. Altogether, we refer to this process as combinatorial empirical prompting, and, in our case studies, the approach produced consistent and substantial improvements in LLM performance across constructs and models. Following the baseline prompt, the second most effective approach was the empirical selection of few-shot examples. After a high-performing baseline prompt and set of examples have been identified, we find that additive techniques offer only small and inconsistent improvements.

Overall, we find that an optimal prompt can improve alignment with human classifications, when compared to an ineffective prompt, by as much as 22\% on a low-inference construct and by as much as 65\% on a high-inference construct. While alignment with human classifications alone is not a comprehensive assessment of validity, it remains a foundational source of validity evidence. We demonstrate how this aspect of validity can be meaningfully improved through prompt engineering, a component of LLM-based classification that, unlike model architecture and pre-training, remains fully within researcher control.

\section*{Automatic Construct Identification in Psychology}

Researchers often turn to participants’ own words (in their speech or their writing) as expressions of their “subjective, inner state (\cite{gottschalk_measurement_2022})”. When these states are systematically measured by identifying the presence or strength of constructs within participant text, hypotheses can be quantitatively tested on fundamentally qualitative data sources (\cite{guetzkow_unitizing_1950}). Yet, obtaining precise measurements of psychological constructs from text can be time consuming and challenging. When performed manually, this process is termed coding and typically requires either experts or trained non-experts to 1) come to consensus on construct definitions and their application, 2) individually comb through texts to identify those constructs, and 3) assess and reconcile discrepancies to ensure consistency and reliability (\cite{shaffer_how_2021}).

Given the challenges of expert coding, there is an established line of research using computational techniques to automate the identification of constructs, reaching back to the General Inquirer in the 1960s (\cite{boyd_natural_2021}; \cite{stone_general_1966}). For decades, this often involved the counting of psychologically-relevant terms, through theoretically-derived dictionaries (\cite{tausczik_psychological_2010}). As computational capacity and machine learning methods developed, approaches grew to include supervised classification. Under this approach, researchers complete the full manual coding process but only on a subset of texts. This subset is used to train a machine learning model and then to assess its performance. The supervised approach has been used in the identification of motivational interviewing skills (\cite{atkins_scaling_2014}; \cite{can_it_2016}; \cite{gibson_deep_2016}; \cite{imel_computational_2015}), emotions (\cite{tanana_how_2021}), depression (\cite{jamil_monitoring_2017}), suicidal ideation and attempt (\cite{bejan_improving_2022}), and mental schemas relevant to cognitive behavioral therapy (\cite{burger_natural_2021}), among others. 

More recently, with the public release of pre-trained large language models, psychological researchers have increasingly turned to zero-shot classification, few-shot classification, and fine-tuning (obtaining updated model weights with a small training dataset) to identify constructs. Recent applications include the identification of therapist actions like affirmation and reflective listening in therapy transcripts (\cite{hammerfald_leveraging_2025}), client emotions in therapy transcripts (\cite{lalk_employing_2025}), meta-cognitive thinking in client reflections (\cite{stavropoulos_shadows_2024}), and a variety of mental states and experiences like depression and harm as expressed in social media data (\cite{bunt_validating_2025}; \cite{xu_mental-llm_2024}). 

\subsection*{Validation for Construct Identification}

In both supervised and LLM-based classification, the most common approach to validation is the comparison of human and machine classifications, and the calculation of performance metrics like recall, precision, specificity and F1. This serves as evidence of a kind of criterion validity, assessing how well a measure correlates with an established standard (\cite{messick_validity_1995}), where, in this case, the human classifications are considered the standard. Such validation assumes, of course, that the human classifications have their own evidence of validity (\cite{anglin_addressing_2024}), often provided via a description of the training process, the consensus processes, and inter-rater reliability (as in \cite{stavropoulos_shadows_2024}, p. 7635). In supervised classification, alignment with human classifications is directly optimized for during model training (as weights and features are selected to reduce differences between machine and human classifications). Empirical prompt engineering can involve an analogous process where a prompt is selected to maximize alignment with trusted human classifications (\cite{bunt_validating_2025}). After the prompt is selected, it is validated against a hold-out testing sample, not used to select the prompt in order to avoid overfitting (\cite{anglin_automatic_2025}). 

Off the shelf, LLMs can often easily pass informal assessments of face validity; for example, when prompted to classify texts and explain those classifications, their explanations are cogent and often convincing (\cite{wang_evaluating_2023}). Yet, the classifications may not reflect expert understandings of the construct, particularly when those constructs are field-specific and latent (\cite{stavropoulos_shadows_2024}). Because LLMs are trained on limited academic literature, given that literature is often protected by a paywall (\cite{cook_pushing_2025}), phrases like “negative core beliefs” may have few occurrences in their pre-training data, and what occurrences there are likely reflect informal understandings. Validity challenges are compounded when 1) constructs are latent and must be inferred from linguistic characteristics that are subtle, differentially expressed across individuals, and must be weighed against one another (\cite{shadish_experimental_2001}); and 2) construct definitions are contested or overlapping with other constructs, such that two experts may disagree on definitions and applications (\cite{shaffer_how_2021}). In these cases, it is particularly unclear which internal representations are activated by the text, or why, and it is unclear to what degree we should trust that a model’s explanations reflect the decision process (\cite{ye_unreliability_2022}). For this reason, validation remains an essential part of the classification process.

What constitutes a comprehensive validation process for LLM-based classifications is still an on-going area of discussion. \cite{bunt_validating_2025} for example, recommend not only assessing alignment with human classifications, but also assessing the nature of discrepancies and examining whether the language used in the prompt reflects the meaning of the construct of interest (\cite{bunt_validating_2025}). Other validation approaches, drawn from the broader measurement literature, are also available, such as assessing correlations between model classifications and future outcomes (\cite{american_educational_research_association_standards_2013}). To date, however, alignment between model and human classifications remains the standard method of validation (as evidenced by \cite{hammerfald_leveraging_2025}; \cite{lalk_employing_2025}; \cite{stavropoulos_shadows_2024}; \cite{xu_mental-llm_2024}) and a practical approach to prompt selection, offering a straightforward and empirical basis for guiding the model toward a classification scheme that is more likely to closely reflect the construct of interest. 

\section*{Prompting for Construct Identification}

Generative LLMs are designed to generate text continuations. Given text provided by a user – the prompt – they generate reasonable subsequent words by identifying high-probability continuations that reproduce patterns observed in their training data (\cite{hussain_tutorial_2024}). Within the construct identification context, for example, an LLM might identify a high-probability next word for the following prompt: 

\begin{displayquote}
   “In the following text, an individual responds to questions about their past experiences.
Your task is to determine whether the text reflects one or more specific negative core beliefs. Negative core beliefs are generalized or exaggerated judgments people have about themselves, others, or the world. Report your final classification as either \textasciigrave{}\textasciigrave{}\textasciigrave{}yes\textasciigrave{}\textasciigrave{}\textasciigrave{} or \textasciigrave{}\textasciigrave{}\textasciigrave{}no\textasciigrave{}\textasciigrave{}\textasciigrave{} surrounded by triple backticks.” 
\end{displayquote}
\begin{flushright}
    Prompt A
\end{flushright}

Depending on the content of the text, the next most likely word will be either “\textasciigrave{}\textasciigrave{}\textasciigrave{}yes\textasciigrave{}\textasciigrave{}\textasciigrave{}” or “\textasciigrave{}\textasciigrave{}\textasciigrave{}no\textasciigrave{}\textasciigrave{}\textasciigrave{}.”

Given the model aims to maximize the probability of additional terms, conditional on the previous terms, adjusting the previous terms (the prompt) will change the probability of a given classification. 

\subsection*{Baseline Prompting}
The example prompt above provides context (defining negative core beliefs), and designates both the desired output (e.g., tasking the LLM with determining if negative core beliefs are evident in some text) and format (\textasciigrave{}\textasciigrave{}\textasciigrave{}Yes\textasciigrave{}\textasciigrave{}\textasciigrave{} or \textasciigrave{}\textasciigrave{}\textasciigrave{}No\textasciigrave{}\textasciigrave{}\textasciigrave{}). Together, we refer to these components as a baseline prompt. Given a baseline prompt is written in natural language (as opposed to artificial language or computer code), often a reasonable starting place is the same set of instructions which would be provided to a human coder (\cite{gilardi_chatgpt_2023}; \cite{michelmann_large_2025}). 

Intuition and theory suggest that baseline prompts are most effective when instructions are unambiguous and specific (\cite{chen2023unleashing}; \cite{lo2023art}). Other practical (though not always empirical) advice includes ensuring that the label provided for the text classification task (here, negative core beliefs) matches common usage of language so the LLM can locate the definition of the construct within the pre-training data (\cite{gao_making_2021}), incorporating bullet points to increase clarity (\cite{mishra_reframing_2022}), and preferencing positive rather than negative statements (\cite{battle_unreasonable_2024}). Other suggestions remain contested. For example, while \cite{reynolds_prompt_2021} argue that redundancy in prompting can improve performance, \cite{mishra_reframing_2022} recommend prioritizing conciseness to avoid confusion or overload.

Unfortunately, the best performing baseline prompt is not always intuitive; “seemingly worthless prompt modifications” can have meaningful performance impacts (\cite{battle_unreasonable_2024}, p. 1). Indeed, the best performing prompt may not even be sensible to human readers (\cite{battle_unreasonable_2024}). This is one reason why \textit{automatic prompt engineering} can be a useful approach. Automatic prompt engineering supplements hand-designed prompts by first using the LLM to iteratively generate a set of alternative prompts given a seed prompt; and second, selecting the highest performing prompt among the alternatives and using this as the next seed (\cite{zhou2022}). Previous tests of this approach have indicated mixed performance, with the automatically engineered prompt sometimes performing worse than the default but often performing comparable or better (\cite{zhou2022}). 

\subsection*{The Use of Examples}

Eliciting valid classifications from an LLM without examples or demonstrations – i.e., via zero-shot classification – can be difficult. As an alternative, one-shot classification not only provides a task but also a demonstration of it (\cite{brown_language_2020}; \cite{weber_evaluation_2023}). For example, to use one-shot, we would extend Prompt A with one example of a written response from a participant that has been labeled with either “Yes” or “No” (indicating a negative core belief, or not). In doing so, we provide the model with a clear demonstration of the task at hand. Few-shot classification extends the principle of one-shot classification by increasing the number of examples and thus providing additional context-specific knowledge on the task (\cite{brown_language_2020}). The number of demonstrations in few-shot learning is up to the user and can range between as few as two and as many as dozens (\cite{brown_language_2020}). 

\subsection*{Additive Prompting Techniques}

Recent research into prompt engineering has often focused on what we term here \textit{additive} prompt engineering techniques. These are broadly applicable techniques that may be applied to any baseline prompt. In what follows, we examine three additive techniques, ordered by the degree of adaptation they require, from minimal to substantial. 

\subsubsection*{Persona Pattern}

A persona pattern is a short phrase, commonly added to the beginning of the prompt, which provides the LLM with either a role to play or a characteristic to embody (\cite{white_prompt_2023}). In theory, appending a persona pattern works because it draws on a kind of “cultural consciousness” associated with the role (\cite{reynolds_prompt_2021}, p. 4). In calling on the therapist or psychologist role, for example, we hope that the model is also activating patterns associated with negative core beliefs in the psychology context. 

\subsubsection*{Chain-of-Thought Prompting}

Zero-shot chain-of-thought prompting involves adding a simple phrase such as, “Let’s think this through step by step” to the end of a prompt (\cite{kojima_large_2022}). Despite the simplicity of the approach, this phrase has improved LLM performance on a range of benchmark tasks, including arithmetic and symbolic reasoning (\cite{golchin_grading_2025}; \cite{kojima_large_2022}). In few-shot chain-of-thought prompting, reasoning is not just encouraged (via a phrase) but demonstrated with examples (\cite{wei_chain--thought_2022}). Within the context of text classification, reasoning exemplars highlight the aspects of text that need to be attended to and how they relate to the definition of the construct. 

\subsubsection*{Explanations}

Explanations are justifications paired with few-shot examples (\cite{lampinen_can_2022}). The key difference between few-shot chain-of-thought prompting and explanations is the order: with chain-of-thought, the justification occurs before the classification. With explanations, the justification is provided after. Perhaps surprisingly, this can make a difference. These two approaches likely differ in the weight provided to the justification compared to the exemplar answers. Due to recency bias, whichever occurs last is likely to be given the greater weight by the model (\cite{liu_pre-train_2023}). Thus, explanation prompting likely weights the justification more heavily than the example-answer pair. 

\section*{Contribution}

Unfortunately, prior research indicates that none of the above additive prompting approaches can be counted on to reliably improve performance. For example, while chain-of-thought prompting has at times demonstrated significant improvements in model performance (\cite{wei_chain--thought_2022}), in other studies, it has demonstrated an adverse effect (\cite{anil_palm_2023}). Further, model performance can depend on seemingly minute changes in prompt wording, exemplar selection, and even exemplar order (\cite{sahoo_systematic_2024}; \cite{zhao_calibrate_2021}). Here, we address the seemingly erratic nature of LLM prompting in two ways: first, by searching for generalizable patterns across three constructs and two models, and second, by formulating a systematic, theoretically and empirically grounded prompt engineering protocol that can be used across diverse text classification tasks.

\section*{Case Study Experiments and Data}

LLMs tend to be reliable text classifiers for low-inference concepts like sentiment or discrete emotions (\cite{rathje2024gpt}). However, high-inference concepts that have a complex, field-specific meaning – what \cite{ziems_can_2024} refer to as expert taxonomies – pose a more difficult challenge. Here, we design our experiments to capture a spectrum of scenarios from low complexity to high complexity, both with and without expert taxonomies. 

\subsection*{Low Complexity – Emotion Identification}

We generate a low-inference task by drawing on a dataset of Reddit posts labeled with the emotions they express (\cite{demszky_goemotions_2020}). The original dataset includes over 58,000 Reddit posts, labeled with 26 emotions by 3-5 annotators each (described as native English speakers from India). Interrater reliability in that dataset ranges from a Cohen’s Kappa of 0.144 (for nervousness) to 0.749 (for gratitude). For the purposes of this study, we focus on the emotion with the highest interrater reliability – gratitude – and limit our analysis to a stratified random sample of 800 texts (a sample size comparable to the other case studies), half labeled as expressing gratitude. Gratitude is defined by the dataset creators as “a feeling of thankfulness and appreciation” (\cite{demszky_goemotions_2020}, p. 12). 

\subsection*{Higher Complexity – Positive Meaning Making and Negative Core Beliefs}

Many psychological constructs are complex (with multi-component definitions), high inference (requiring interpretation), and overlapping (difficult to distinguish from related constructs). To assess the effectiveness of prompt-engineering techniques in these more challenging scenarios, we experiment with two additional constructs: positive meaning making and negative core beliefs. We define positive meaning making as occurring when a person reports a positive change or outcome resulting from a negative event, trauma, or stressor in their life (\cite{boals_use_2012}). We define negative core beliefs as broad, generalized, or exaggerated negative judgments people have about themselves, others, or the world (\cite{beck_cognitive-behavioral_2011}). We identify instances of positive meaning making and negative core beliefs in texts produced within expressive writing interventions (\cite{reinhold_effects_2018}), in which participants engaged in independent, reflective writing over a brief period. 

We draw from recently completed studies of expressive writing interventions focused on young (ages 21 to 30) biological or trans- women who have been previously diagnosed with a mental health condition. Our sample includes 416 unique individuals who completed between one and three expressive writing tasks. The responses to these tasks (hereafter, texts) serve as the unit of analysis; 802 texts were labeled for meaning making and 793 texts were labeled for negative core beliefs. Table \ref{table:1} provides information on the distribution of construct prevalence. Meaning making was independently coded by undergraduate psychology students (estimated Cohen’s Kappa = 0.70), with a graduate student and faculty member resolving disagreements. Negative core beliefs were independently coded by a psychology graduate student and faculty member, with an estimated Cohen’s Kappa of 0.82. Although negative core beliefs has higher interrater reliability than gratitude, we nonetheless conceptualize it as a more difficult, higher-inference construct. We suspect the relatively high reliability reflects differences in coder expertise (negative core beliefs was coded by graduate- and post-graduate-level coders), training, and consensus procedures, in addition to construct complexity.

\subsection*{Strengths and Limitations of the Case Studies}

Our case studies offer three key strengths. First, they span a range of task complexity, from low-inference classification (e.g., emotion recognition) to higher-inference construct classification (e.g., positive meaning making and negative core beliefs), allowing us to assess prompting techniques across levels of task difficulty. Second, the higher-inference concepts represent constructs that require LLMs to use “unique reasoning paths not typically encountered…during their pre-training phase” (\cite{lee_applying_2024}). Presumably, prompt engineering is of even greater importance for such constructs. Third, the classified texts reflect real-world writing, including spelling errors, irrelevant content, and content on which two experts can reasonably disagree. 

On the other hand, our data present generalizability challenges. Notably, the expressive writing texts for meaning making and negative core beliefs were generated only by women. This would be of particular concern if our goal was to assess the performance of LLMs or to suggest the best prompt for these constructs. Instead, our goal is to assess the usefulness of prompt engineering approaches. Thus, the limited diversity of our sample will only threaten our results if there is a reason to believe, for example, that persona prompts work better or worse for women’s writing compared to men. 

Finally, because performance metrics are only strong evidence of validity when the human benchmark has validity evidence itself, it is worth considering the validity of our three constructs beyond interrater reliability (as multiple coders may code according to the same misunderstanding of a construct). Because we draw on an open-source dataset for gratitude, we have limited validity evidence for this construct beyond interrater reliability. For meaning making and negative core beliefs we offer the following. First, the codebooks describing these constructs were based on an established theoretical literature base (e.g., \cite{beck_cognitive-behavioral_2011}; \cite{boals_use_2012}; \cite{boals_coping_2011}; \cite{dozois_core_2015}; \cite{jorovat_core_2025}; \cite{osmo_negative_2018}; \cite{wenzel2012}). Second, annotators for these constructs had a background in psychology (either at the undergraduate or graduate level), with undergraduates coding only after demonstrating sufficient alignment with supervising graduate students. Finally, one author of this paper, with clinical experience, conducted periodic checks of classifications and participated in consensus conversations.

\section*{Methods}

We split our data such that 33\% of individuals are assigned the training dataset, 33\% to the development dataset, and 33\% to the testing dataset. We chose this distribution for two reasons. First, compared to training a supervised learning classifier from scratch, our training needs are limited. We use training data solely to select the baseline prompts and to provide and select examples for few-shot learning. Second, given the primary purpose of this manuscript is to compare prompting approaches, we devote a substantial portion of our labeled data to this purpose using the development dataset. Still, as a demonstration of best practice, we retain a testing dataset that is large enough to credibly assess the performance of the final selected model. Table \ref{table:1} provides sample sizes for each of our constructs and data splits. 

\begin{table}
\caption{\textit{Training, Development, and Testing Sample Sizes}}
\begin{center}
\begin{tabular}{l c c c c c c c c}
    \hline
    \multirow{2}{*}{Concept} & \multirow{2}{*}{\shortstack{Proportion \\ with Concept}} & \multicolumn{2}{c}{Train} & \multicolumn{2}{c}{Dev} & \multicolumn{2}{c}{Test} & Total\\ &
         & Yes & No & Yes & No & Yes & No & All \\
        \hline
        Gratitude & 0.50 & 122 & 144 & 142 & 124 & 136 & 132 & 800 \\
        Meaning Making & 0.26 & 71 & 174 & 72 & 192 & 67 & 226 & 802\\
        Negative Core Beliefs & 0.25 & 66 & 181 & 65 & 204 & 69 & 208 & 793\\
        \hline
\end{tabular}
\end{center}
\textit{Note.} We purposefully limit the sample size of the gratitude dataset to be more representative of the size of datasets commonly included in psychological studies. 
\label{table:1}
\end{table}

\subsection*{Metrics}

When assessing the performance of our prompts, we use five criteria: accuracy, specificity, precision, recall, and F1. Each metric is based on the relationship between true positives (i.e., instances where the LLM correctly identifies an instance of the construct), false positives (incorrectly identifying the construct), false negatives (incorrectly identifying the absence of the construct), and true negatives (correctly identifying the absence of the construct). Here, “true” and “false” are defined relative to the human label. (Though the use of these terms is something of a shorthand, given we know that human labels bring their own error, a fact exemplified by our assessment of interrater reliability). 

Accuracy measures the overall proportion of correct classifications (true positives and true negatives). Specificity measures the proportion of human negatives that the model also identified as negative. Recall, or true positive rate, measures the proportion of human positives that were successfully identified by the model. Precision measures the proportion of positive identifications by the model that the human also identified as positive. F1 balances precision and recall into a single metric, providing a harmonic mean that is particularly useful when a construct is rare. Formally, these are defined as:

\begin{gather*}
    Accuracy = \frac{TN + TP}{TN+TP+FN+FP}, \\
    Specificity = \frac{TN}{TN+FP},\\
    Recall = \frac{TP}{TP+FN}, \\
    Precision = \frac{TP}{TP+FP}, \\
    F1 = \frac{2 \cdot Precision \cdot Recall}{Precision + Recall}
\end{gather*}

\subsection*{Models}

We perform our experiments using one closed-source, proprietary LLM from OpenAI (\courier{gpt-5.4-2026-03-05}; \courier{gpt-5.4}) and one open-source LLM from Meta (\courier{llama3.3}). We employ \courier{gpt-5.4} programmatically through the \courier{openai} Python library and call \courier{llama3.3} through Ollama. We run \courier{llama3.3} within our university’s high-performance computing (HPC) environment where we requested 3 GPUs and 128 GB of memory on a single node. We installed \courier{llama3.3:latest} from the Ollama Library which contains 70.6B parameters and is set to 4-bit quantization. 

For \courier{llama3.3}, we set temperature to zero (reducing randomness in token selection) and for \courier{gpt-5.4} we set reasoning to None. In Appendix \ref{table:app-a}, we compare performance across a combination of OpenAI models, temperatures, and reasoning settings on a random sample of 70 texts from our training dataset. The results suggest that 1) \courier{gpt-5.4} outperforms earlier models; 2) a higher level of reasoning does not consistently increase performance but does increase cost; and 3) likewise, a higher temperature does not consistently increase performance but does consistently decrease reliability. 

In this article, we are agnostic to the choice of LLM platform and assess generalizability across platforms. However, there are ethical and privacy considerations in the choice of LLM provider, leading researchers to commonly prefer the use of open-source and open-access LLMs when feasible (\cite{hussain_tutorial_2024}). Closed-access models, like those from OpenAI, require that data be shared with the provider. At the time of this writing, texts provided to OpenAI through the Python API are not retained by OpenAI long-term. Instead, OpenAI states that they ‘‘securely retain API inputs and outputs for up to 30 days to provide the services and to identify abuse. After 30 days, API inputs and outputs are removed from our systems, unless [OpenAI is] legally required to retain them’’ (\cite{openai_enterprise_2025}). Still, this privacy policy brings concerns when it comes to personally identifying information (which, in our case, was removed). On the other hand, because open-access models can be run locally (ensuring that data never leave the user’s system), they provide stronger protections for participant privacy. 

\subsection*{Baseline Prompts and Examples}

\subsubsection*{Combinatorial Empirical Prompting}

We define combinatorial empirical prompting as systematically recombining codebook-derived components and empirically evaluating the resulting prompts. In line with best practices for codebook development, our baseline prompts incorporate four components: contextual information, a task instruction, a construct definition, and guidance on inclusion/exclusion criteria (\cite{macqueen_codebook_1998}). Because each of these components may influence LLM performance, we manually generated multiple variants of each, including many potential inclusion/exclusion criteria drawn from a codebook used by human coders. To ensure that guidance criteria were comprehensive, we also observed texts in the training dataset, looking for rules that may be commonly understood by the coders but not explicit in the codebook. Where such rules were identified (e.g., that past negative core beliefs, described by the author as something they no longer believe, should be coded as no), we appended them as potential inclusion/exclusion criteria. We intentionally avoided LLM-generated variants here as automated prompts are built into the automated prompt engineering process discussed later. Because the construct label itself may influence model performance (particularly as academic terms may be unfamiliar in the pre-training data), before generating prompts, we piloted multiple names for the construct, including: “negative assumptions” and “general negative beliefs” as alternatives to “negative core beliefs” and “positive benefit” and “reappraisal” as alternatives to meaning making. Both “negative core beliefs” and “meaning making” outperformed our alternatives. Thus, these were the terms we used when generating prompting components. See Table \ref{table:2} for examples of each prompt component. 

Using the following procedure, we generated 50 prompt combinations for each construct:

\begin{enumerate}
    \item Randomly select one context description variant (from options including an empty string, in case the context description is not necessary or useful).
    \item Randomly select one task description variant.
    \item Randomly select one construct definition variant.
    \item Randomly draw a number from 0 to n. If the number is greater than 0, append that number of randomly selected inclusion/exclusion criteria to the definition.
    \item Append a final formatting description.
\end{enumerate}

We repeated this process 50 times to produce 50 codebook-guided, human-generated prompts stored in an Excel file with a unique ID, which can later be imported into \courier{python}. Each variant was evaluated on our training dataset, with results exported to a CSV file.

Then, we identified the prompts yielding the highest and lowest F1 scores for each construct-model. Together, the top and bottom prompts serve two purposes. First, the performance gap between the two speaks to the extent to which reasonable variations in baseline prompt wording can influence model output—and to the utility of an empirical prompt selection process. Second, the lowest-performing prompt serves as a test of whether few-shot examples, automatic prompt engineering, and additive techniques can fix a prompt that, on its own, does not produce classifications aligned with human codes. 

\begin{table}
\caption{\textit{Baseline Prompt Components with Example from Negative Core Beliefs}}
\begin{center}
\begin{tabular}{l c c p{9cm}}
    \hline
    Component & \shortstack{Number\\Generated for\\Negative\\Core Beliefs} & \shortstack{Number\\Sampled} & Example\\
        \hline
        Context & 4 & 1 & As part of a therapeutic program, clients have been asked to write responses to questions about their identities and their childhood, including positive and negative experiences. You are about to read a writing sample in response to one of those questions.\\
        Task & 4 & 1 & Your task is to determine whether the text either strongly implies or explicitly states a negative core belief about the self or others currently held by the author.\\
        Definition & 4 & 1 & Negative core beliefs are broad, generalized, or exaggerated negative beliefs that people hold about themselves or others.\\
        Guidance & 15 & 0-15 & To determine if there is sufficient evidence of a negative core belief, you might try to complete one of these sentences: “I am \underline{\hspace{1cm}},” “People are \underline{\hspace{1cm}},” or “People will \underline{\hspace{1cm}}”.\\
        Formatting & 1 & 1 & Report your final classification as either \textasciigrave{}\textasciigrave{}\textasciigrave{}yes\textasciigrave{}\textasciigrave{}\textasciigrave{} or \textasciigrave{}\textasciigrave{}\textasciigrave{}no\textasciigrave{}\textasciigrave{}\textasciigrave{} surrounded by triple backticks. 
        Here is the text:
        \{TEXT\}
        Remember, report your final classification as either \textasciigrave{}\textasciigrave{}\textasciigrave{}yes\textasciigrave{}\textasciigrave{}\textasciigrave{} or \textasciigrave{}\textasciigrave{}\textasciigrave{}no\textasciigrave{}\textasciigrave{}\textasciigrave{} surrounded by triple backticks.\\
        \hline
\end{tabular}
\end{center}
\textit{Note.} For gratitude, we generate three context, four task, and four definition variants, as well as eight potential guidance criteria. For positive meaning making, we generate four context, four task, and five definition variants, as well as 12 guidance criteria. All prompts include the same formatting instructions. 
\label{table:2}
\end{table}

\subsubsection*{Automatic Prompt Engineering}
To automatically generate prompts, we followed an iterative process beginning with a human-generated baseline prompt (either the top- or bottom-performing baseline prompt) as the seed. At each iteration, we provided a seed prompt within the following meta-instruction:
\begin{displayquote}
   “Generate a variation of the following prompt while keeping the output format. You can add important information or remove unnecessary information. Instruction:\textbackslash n Output only the new instruction.” 
\end{displayquote}
\begin{flushright}
    Prompt B
\end{flushright}

We had the relevant model generate five new prompts per iteration in response to this meta-prompt. Each prompt was evaluated on the training dataset, and the prompt yielding the highest F1 score was selected as the new seed. This process was repeated for five generations, and we selected the highest performing prompt across all iterations. When using \courier{llama3.3}, temperature for the classification task was set to 0 while temperature for the prompt generation was set to 1. 

\subsubsection*{Few-Shot Example Selection}

To create a pool of potential few-shot examples, we siphoned off 30 examples from the training dataset: 15 randomly selected positive examples and 15 randomly selected negative examples. To produce random combinations of these examples, we first randomly selected a number from 4 to 10; this defines the number of examples n. Then, we randomly select n examples from the pool without replacement. We repeated this process 50 times to yield 50 distinct few-shot combinations. We evaluated each combination using the remaining training data (i.e., texts not included in the example pool) and identify the combination that achieves the highest F1 score, for both the top and bottom baseline prompts.

\subsection*{Additive Techniques}

\subsubsection*{Persona}
We generated three potential personas for each construct. For gratitude, we tested: 
\begin{enumerate}
    \item You are a language model trained to detect emotions in user text. 
    \item You are an expert annotator.
    \item You are an expert annotator trained on a detailed coding manual for identifying gratitude.
\end{enumerate}

For negative core beliefs and meaning making, we tested: 
\begin{enumerate}
    \item You are a clinical psychologist trained in cognitive behavioral therapy.
    \item You are an expert annotator trained on a detailed coding manual for identifying \{CONSTRUCT\}.
    \item You are a mental health counselor who carefully reads their clients' writing to inform the focus of the next session.
\end{enumerate}

For each construct, we selected the persona that generated the highest performance in the training dataset when prepended to the top and bottom baseline prompts. 

\subsubsection*{Chain-of-Thought}

To incorporate chain-of-thought prompting into a zero-shot classification task, we appended the following to the baseline prompts: “First, state what you notice in the text as it relates to the definition. Then, report your final answer as either \textasciigrave{}\textasciigrave{}\textasciigrave{}yes\textasciigrave{}\textasciigrave{}\textasciigrave{} or \textasciigrave{}\textasciigrave{}\textasciigrave{}no\textasciigrave{}\textasciigrave{}\textasciigrave{}.” For few-shot classification, we included chain-of-thought reasoning for each selected example. Given potential sensitivity to reasoning phrasing, we generated two alternative explanations per example—one written by a human and one generated by \courier{gpt-5.4}. Rather than exhaustively evaluating all combinations, we constructed a subset of prompt variants: for each prompt type (top-performing and bottom-performing baseline prompts), we included the all-human and all-model reasoning configurations, along with three additional combinations sampled from the remaining possibilities. This yields five chain-of-thought variants each. These were evaluated on the training set and we selected the highest-F1 combination separately for the top-performing and bottom-performing baseline prompts.

\subsubsection*{Explanations}

To incorporate explanations into our prompts, we used the following prompt template: 

\{Baseline Prompt\} 

Text: \{Example\} 

Answer: \{Answer\} 

Explanation: \{Explanation\}

We followed the same process as for few-shot chain-of-thought, generating five prompt variants for the top-performing prompt (consisting of combinations of explanations) and five prompt variants for the bottom-performing prompt. For each, we retained the highest F1 prompt.

\subsection*{Evaluation}

To assess the extent to which each technique improves alignment between human and LLM classifications, we compare F1 scores calculated in the development dataset. After identifying the technique and model which resulted in the best performance for each construct, we report accuracy, specificity, recall, precision, and F1 for this final approach as estimated in the testing dataset. Both sets of metrics are accompanied by non-parametric hierarchical bootstrapped standard errors that take into account the nested nature of the data (texts nested within individuals), calculated as follows: 1) we sampled \textit{n} participants with replacement (where n is the number of unique participants in the development or testing dataset); 2) for each sampled participant, we sampled with replacement from texts produced by that individual, creating the bootstrap sample; 3) we computed the metric for each of 1,000 bootstrap samples; 4) we calculated the standard deviation of the bootstrap distribution to generate standard errors; 5) we identified the 2.5th and 97.5th percentiles of the bootstrap distribution to generate 95\% confidence intervals.

\section*{Results}

Tables \ref{table:3} through 8 present F1 scores as estimated in the development dataset for each of the three constructs: gratitude (Tables \ref{table:3}-\ref{table:4}), meaning making (Tables \ref{table:5}-\ref{table:6}), and negative core beliefs (Tables \ref{table:7}-\ref{table:8}); \courier{gpt-5.4} results are in the odd-numbered tables while \courier{llama3.3} results are in the even-numbered tables. In each, the left panel provides the performance of the bottom prompt, while the right panel provides the performance of the top prompt, both identified using combinatorial empirical prompting. Alongside the baseline prompts, we present the performance of prompts with additional automatic prompt engineering (with the top/bottom as the seed prompt), personas, chain-of-thought, explanations, and few-shot examples. Asterisks indicate statistically significant differences between a given baseline prompt (highlighted in gray) and these additional techniques. 

\begin{table}
\caption{\textit{The Performance of Prompting Techniques for Gratitude on gpt-5.4, as Assessed in the Development Dataset}}
\begin{center}
\begin{tabular}{l c c c c}
    \hline
     &  \multicolumn{2}{c}{Bottom Baseline} & \multicolumn{2}{c}{Top Baseline}\\
     & Zero-Shot & Few-Shot & Zero-Shot & Few-Shot\\
        \hline
        Baseline & \cellcolor[gray]{0.90}0.77 & 0.86** & \cellcolor[gray]{0.90}0.87 & \textbf{0.93**}\\
        & \cellcolor[gray]{0.90}(0.03) & (0.02) & \cellcolor[gray]{0.90}(0.02) & \textbf{(0.02)}\\
        Automatic Prompt Engineering & 0.80 & 0.87*** & 0.90 & 0.92*\\
        & (0.03) & (0.02) & (0.02) & (0.02)\\
        Persona & 0.78 & 0.86** & 0.86 & 0.92*\\
        & (0.03) & (0.02) & (0.02) & (0.02)\\
        Chain-of-Thought & 0.78 & 0.81 & 0.86 & 0.89\\
        & (0.03) & (0.03) & (0.02) & (0.02)\\
        Explanations & & 0.84* & & 0.91\\
        & & (0.03) & & (0.02)\\
        \hline
\end{tabular}
\end{center}
\textit{Note.} The baseline prompt is highlighted in grey. The highest performing prompt is emboldened. Standard errors are in parentheses and estimated via bootstrapping. 
\newline
*$p < 0.10$, **$p < 0.05$, ***$p < 0.01$. 
\label{table:3}
\end{table}

\begin{table}
\caption{\textit{The Performance of Prompting Techniques for Gratitude on llama3.3, as Assessed in the Development Dataset}}
\begin{center}
\begin{tabular}{l c c c c}
    \hline
     &  \multicolumn{2}{c}{Bottom Baseline} & \multicolumn{2}{c}{Top Baseline}\\
     & Zero-Shot & Few-Shot & Zero-Shot & Few-Shot\\
        \hline
        Baseline & \cellcolor[gray]{0.90}0.76 & 0.82 & \cellcolor[gray]{0.90}0.88 & 0.91\\
         & \cellcolor[gray]{0.90}(0.03) & (0.03) & \cellcolor[gray]{0.90}(0.02) & (0.02)\\
        Automatic Prompt Engineering & 0.85** & 0.87*** & 0.87 & 0.89\\
         & (0.02) & (0.02) & (0.02) & (0.02)\\
        Persona & 0.74 & 0.82 & 0.87 & \textbf{0.92}\\
         & (0.03) & (0.03) & (0.02) & \textbf{(0.02)}\\
        Chain-of-Thought & 0.72 & 0.80 & 0.86 & 0.89\\
         & (0.04) & (0.03) & (0.02) & (0.02)\\
        Explanations &  & 0.83* &  & 0.91\\
         &  & (0.03) &  & (0.02)\\
        \hline
\end{tabular}
\end{center}
\textit{Note.} The baseline prompt is highlighted in grey. The highest performing prompt is emboldened. Standard errors are in parentheses and estimated via bootstrapping. 
\newline
*$p < 0.10$, **$p < 0.05$, ***$p < 0.01$. 
\label{table:4}
\end{table}

\begin{table}
\caption{\textit{The Performance of Prompting Techniques for Meaning Making on gpt-5.4, as Assessed in the Development Dataset}}
\begin{center}
\begin{tabular}{l c c c c}
    \hline
     &  \multicolumn{2}{c}{Bottom Baseline} & \multicolumn{2}{c}{Top Baseline}\\
     & Zero-Shot & Few-Shot & Zero-Shot & Few-Shot\\
        \hline
        Baseline & \cellcolor[gray]{0.90}0.81 & 0.87 & \cellcolor[gray]{0.90}0.87 & 0.88\\
         & \cellcolor[gray]{0.90}(0.04) & (0.03) & \cellcolor[gray]{0.90}(0.03) & (0.03)\\
        Automatic Prompt Engineering & 0.78 & 0.82 & 0.87 & \textbf{0.90}\\
         & (0.04) & (0.04) & (0.03) & \textbf{(0.03)}\\
        Persona & 0.80 & 0.85 & 0.88 & 0.88\\
         & (0.04) & (0.03) & (0.03) & (0.03)\\
        Chain-of-Thought & 0.79 & 0.87 & 0.85 & 0.86\\
         & (0.04) & (0.03) & (0.03) & (0.03)\\
        Explanations &  & 0.84 &  & 0.86\\
         &  & (0.04) &  & (0.03)\\
        \hline
\end{tabular}
\end{center}
\textit{Note.} The baseline prompt is highlighted in grey. The highest performing prompt is emboldened. Standard errors are in parentheses and estimated via bootstrapping. 
\newline
*$p < 0.10$, **$p < 0.05$, ***$p < 0.01$. 
\label{table:5}
\end{table}

\begin{table}
\caption{\textit{The Performance of Prompting Techniques for Meaning Making on llama3.3, as Assessed in the Development Dataset}}
\begin{center}
\begin{tabular}{l c c c c}
    \hline
     &  \multicolumn{2}{c}{Bottom Baseline} & \multicolumn{2}{c}{Top Baseline}\\
     & Zero-Shot & Few-Shot & Zero-Shot & Few-Shot\\
        \hline
        Baseline & \cellcolor[gray]{0.90}0.66 & 0.74 & \cellcolor[gray]{0.90}0.80 & \textbf{0.84}\\
         & \cellcolor[gray]{0.90}(0.04) & (0.04) & \cellcolor[gray]{0.90}(0.04) & \textbf{(0.03)}\\
        Automatic Prompt Engineering & 0.75 & 0.79** & 0.77 & 0.80\\
         & (0.04) & (0.04) & (0.04) & (0.04)\\
        Persona & 0.69 & 0.76 & 0.81 & 0.85\\
         & (0.04) & (0.04) & (0.04) & (0.03)\\
        Chain-of-Thought & 0.73 & 0.77* & 0.82 & 0.81\\
         & (0.04) & (0.04) & (0.04) & (0.04)\\
        Explanations &  & 0.78** &  & 0.82\\
         &  & (0.04) &  & (0.04)\\
        \hline
\end{tabular}
\end{center}
\textit{Note.} The baseline prompt is highlighted in grey. The highest performing prompt is emboldened. Standard errors are in parentheses and estimated via bootstrapping. 
\newline
*$p < 0.10$, **$p < 0.05$, ***$p < 0.01$. 
\label{table:6}
\end{table}

\begin{table}
\caption{\textit{The Performance of Prompting Techniques for Negative Core Beliefs on gpt-5.4, as Assessed in the Development Dataset}}
\begin{center}
\begin{tabular}{l c c c c}
    \hline
     &  \multicolumn{2}{c}{Bottom Baseline} & \multicolumn{2}{c}{Top Baseline}\\
     & Zero-Shot & Few-Shot & Zero-Shot & Few-Shot\\
        \hline
        Baseline & \cellcolor[gray]{0.90}0.55 & 0.81*** & \cellcolor[gray]{0.90}0.83 & 0.86\\
         & \cellcolor[gray]{0.90}(0.07) & (0.04) & \cellcolor[gray]{0.90}(0.03) & (0.03)\\
        Automatic Prompt Engineering & 0.66 & 0.83*** & 0.85 & 0.83\\
         & (0.06) & (0.03) & (0.03) & (0.04)\\
        Persona & 0.60 & 0.78*** & 0.82 & 0.89\\
         & (0.07) & (0.04) & (0.03) & (0.03)\\
        Chain-of-Thought & 0.51 & 0.45 & 0.89 & 0.88\\
         & (0.07) & (0.07) & (0.03) & (0.03)\\
        Explanations &  & 0.56 &  & \textbf{0.91*}\\
         &  & (0.07) &  & \textbf{(0.03)}\\
        \hline
\end{tabular}
\end{center}
\textit{Note.} The baseline prompt is highlighted in grey. The highest performing prompt is emboldened. Standard errors are in parentheses and estimated via bootstrapping. 
\newline
*$p < 0.10$, **$p < 0.05$, ***$p < 0.01$. 
\label{table:7}
\end{table}

\begin{table}
\caption{\textit{The Performance of Prompting Techniques for Negative Core Beliefs on llama3.3, as Assessed in the Development Dataset}}
\begin{center}
\begin{tabular}{l c c c c}
    \hline
     &  \multicolumn{2}{c}{Bottom Baseline} & \multicolumn{2}{c}{Top Baseline}\\
     & Zero-Shot & Few-Shot & Zero-Shot & Few-Shot\\
        \hline
        Baseline & \cellcolor[gray]{0.90}0.51 & 0.69*** & \cellcolor[gray]{0.90}0.74 & 0.77\\
         & \cellcolor[gray]{0.90}(0.04) & (0.04) & \cellcolor[gray]{0.90}(0.04) & (0.04)\\
        Automatic Prompt Engineering & 0.60 & 0.76*** & 0.74 & \textbf{0.79}\\
         & (0.04) & (0.04) & (0.05) & \textbf{(0.04)}\\
        Persona & 0.53 & 0.69*** & 0.74 & 0.78\\
         & (0.04) & (0.04) & (0.04) & (0.04)\\
        Chain-of-Thought & 0.60 & 0.76*** & 0.79 & 0.74\\
         & (0.04) & (0.04) & (0.04) & (0.05)\\
        Explanations &  & 0.62* &  & 0.77\\
         &  & (0.04) &  & (0.04)\\
        \hline
\end{tabular}
\end{center}
\textit{Note.} The baseline prompt is highlighted in grey. The highest performing prompt is emboldened. Standard errors are in parentheses and estimated via bootstrapping. 
\newline
*$p < 0.10$, **$p < 0.05$, ***$p < 0.01$. 
\label{table:8}
\end{table}

\subsection*{Combinatorial Empirical Prompting}

Across Tables \ref{table:3}–\ref{table:8}, combinatorial empirical prompting produced substantial differences in performance between the highest- and lowest-performing baseline prompts in the development dataset. This difference in F1 reached as high as 0.28 for classifications of negative core beliefs with \courier{gpt-5.4} (increasing F1 from 0.55 to 0.83). In contrast, the smallest difference was a 0.06 increase in F1 for meaning making on \courier{gpt-5.4}, though even this gap is likely still practically meaningful. With the exception of meaning making on \courier{gpt-5.4}, the difference between the top and bottom prompts was statistically significant across all construct-model combinations. 

\begin{figure}[h!]
    \centering
    \caption{\textit{Histogram of Zero-Shot Baseline F1 Scores from Training Dataset, with Vertical Lines Indicating Performance in Development Set of Bottom, Median, and Top Combinatorial Prompts}}
    \includegraphics[width=\textwidth]{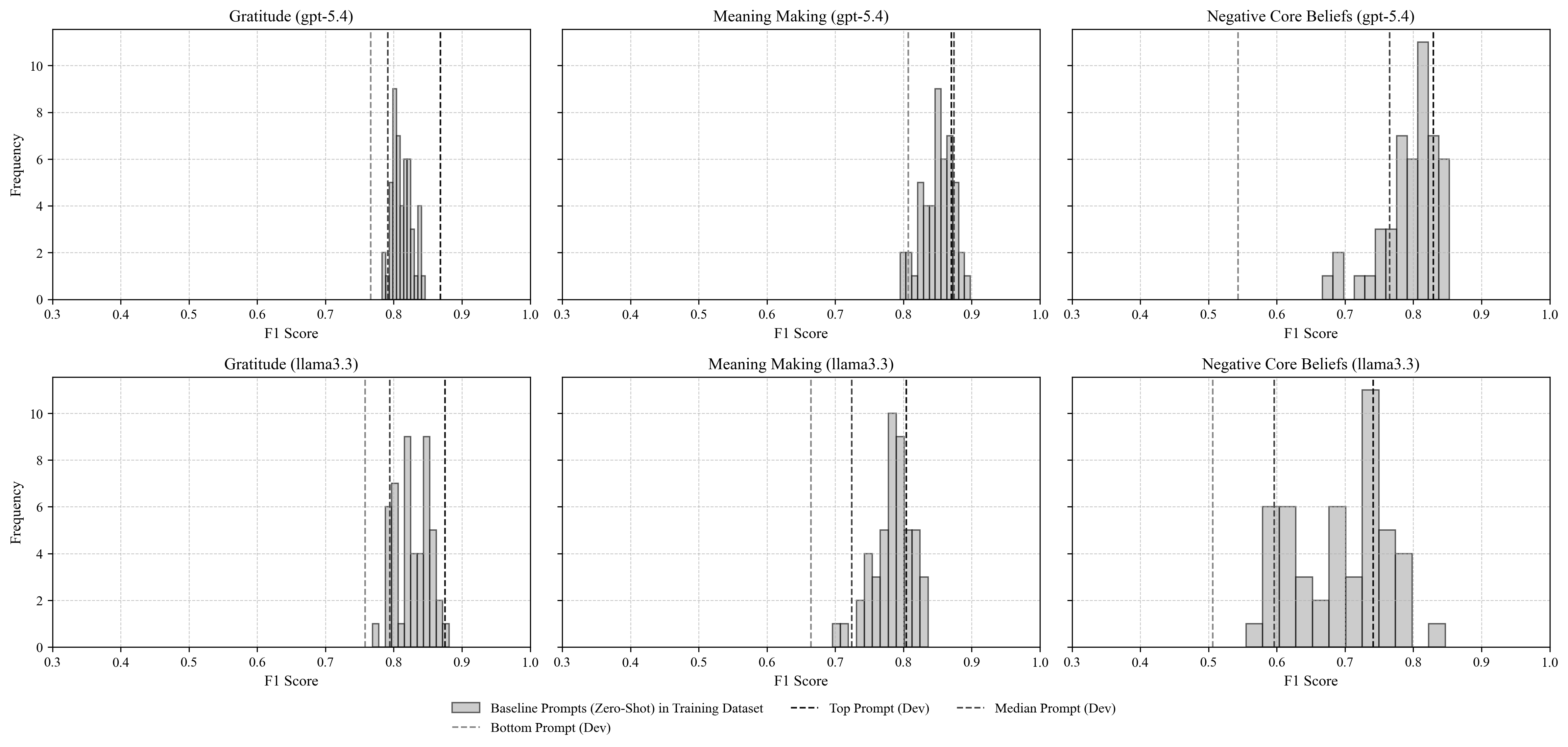}
    \label{fig:a}
\end{figure}

Prompt composition appears to matter most when constructs are difficult to classify. In Figure \ref{fig:a}, we plot the distribution of performance across the 50 baseline prompts in the training dataset. Negative core beliefs has the widest and lowest distribution; gratitude has the tightest and highest. Also in Figure \ref{fig:a}, we overlay development dataset performance for the bottom, top, and median prompts on the histogram, allowing us to explore 1) the added value of the top prompt compared to the median prompt, and 2) the extent to which combinatorial prompting results in prompts that are overfit to the training data. Compared to the median prompt, the top prompt often still demonstrates substantial improvements (in 5 out of 6 cases; all but meaning making on \courier{gpt-5.4}), though rarely statistically significant (in 2 out of 6 cases; gratitude on \courier{gpt-5.4} and \courier{llama3.3}). 

Second, compared to the training dataset, the development dataset does not result in smaller differences in performances between the top and bottom prompts. If it did, this would suggest either a regression to the mean or that the top prompt is overly tailored to the examples in the training dataset. Neither of these patterns appear to be the case; there is only one plot where the vertical lines are fully contained by the histogram. Instead, in some of the concept-model combinations, development performance for the top prompt is above training performance (e.g., gratitude on \courier{gpt-5.4}) while in others it is below (e.g., negative core beliefs on \courier{llama3.3}), suggesting that, while overfit may not be a concern, uncertainty is. Performance metrics vary noticeably, though not statistically significantly different between datasets, illustrating the challenge of moderate sample sizes in the social sciences. On the other hand, \textit{patterns} in performance generalize from the training datasets to the development datasets (i.e., the top-performing training prompt remains higher-performing in the development dataset).

Finally, in Appendices \ref{table:app-b1}-\ref{table:app-b3}, we explore the mechanisms of combinatorial empirical prompting by assessing the relationship between each potential prompt component and prompt performance in the training dataset, focusing on \courier{gpt-5.4}. We estimate, within each construct, the mean difference in training dataset performance among prompts that include and exclude a given component. For example, prompts that include the guidance, “Negative statements that apply to either specific individuals or to ‘some people’ do not imply negative core beliefs. Negative core beliefs must be generalized to many people, or a large group of people”, have a recall that is 0.13 lower on average but a precision that is 0.11 higher. This pattern of components either increasing recall or precision, with decreases in the corollary that are similar in magnitude, is very common: across 71 tested components for the three constructs, only three components increased both precision and recall, and only by 0.01. This suggests that combinatorial empirical prompting improves performance not by identifying strictly beneficial components (i.e., a prompt that includes all of the best components), but by optimizing tradeoffs in recall-oriented and precision-oriented prompt components.

\subsection*{Few-Shot Examples}

Following combinatorial empirical prompting, few-shot examples are the greatest driver of model improvement. In all six construct–model combinations, the addition of few-shot examples to the bottom baseline prompt increases the F1 score, and in three of these combinations, this difference is statistically significant. At a maximum, few-shot prompts generated an increase in F1 of 0.26 in the development dataset (from 0.55 to 0.81 for negative core beliefs on \courier{gpt-5.4}). The addition of few-shot examples to the top baseline prompt also increased the F1 score in all construct–model combinations, albeit to a smaller extent, with a maximum increase of 0.06 (from 0.87 to 0.93 for gratitude on \courier{gpt-5.4}). This is the only combination where the difference between the top baseline zero-shot prompt and the top baseline few-shot prompt is statistically significant.

\begin{figure}[h!]
    \centering
    \caption{\textit{Histogram of Few-Shot Baseline F1 Scores from Training Dataset, with Vertical Lines Indicating Zero-Shot Performance in Training and Development Sets}}
    \includegraphics[width=\textwidth]{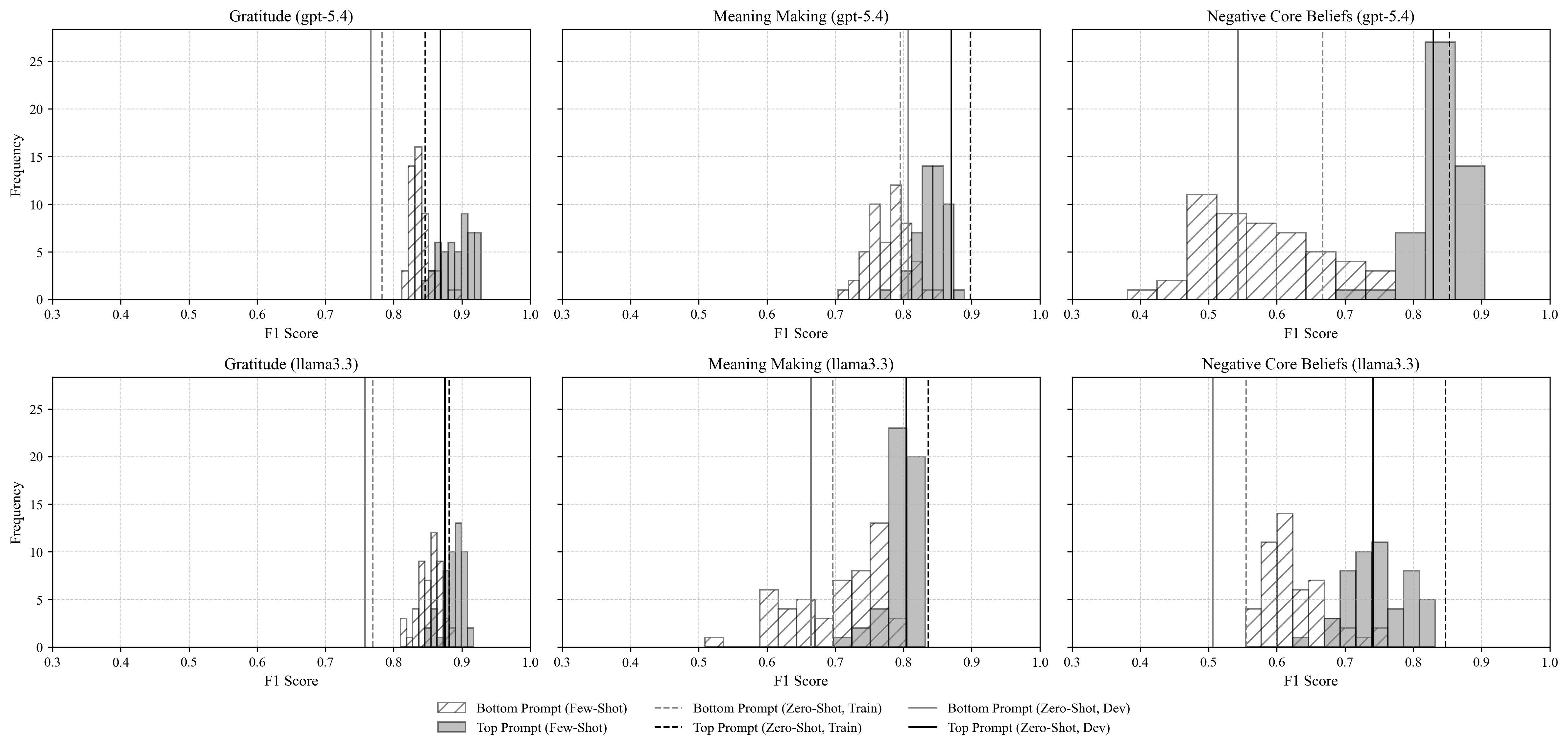}
    \label{fig:b}
\end{figure}

Notably, the specific combination of examples often plays a critical role in determining performance. Figure \ref{fig:b} presents histograms of F1 scores by few-shot combination within the training dataset for both the bottom and top prompts. Dashed vertical lines indicate zero-shot performance in the training dataset, while solid lines indicate zero-shot performance in the development dataset. Many few-shot combinations reduce performance, particularly for high-inference constructs. Further, once again, distributions are tightest for gratitude and widest for negative core beliefs. 

\subsection*{Automatic Prompt Engineering}

After combinatorial empirical prompting and few-shot example selection, automatic prompt engineering is the third most effective technique in our case studies. Results in Tables \ref{table:3}-\ref{table:8} demonstrate that an automatically generated prompt often improves upon an ineffective prompt, increasing performance for five of six construct-model combinations (all but meaning making on \courier{gpt-5.4}). However, the increases are only statistically significant in one case (gratitude on \courier{llama3.3}). Results for the top baseline prompts are smaller and more variable, including three cases where automatic prompt engineering reduced performance, suggesting that automatic prompt engineering is most useful when the initial baseline prompt is weak. It is also important to note that the best-performing automatically generated prompt was rarely produced in the final generation (see Appendix \ref{fig:app-c}). For the top seed prompts, maximum performance was often reached by the first or second generation. For the bottom seed prompts, maximum performance was often reached by the third generation.

\subsection*{Additive Prompting}

Finally, our additive techniques – persona, chain-of-thought, and explanations – offer only mixed improvements. On the one hand, a couple of our best performing prompts within a construct-model incorporate additive techniques. For example, the best performing prompt for negative core beliefs with \courier{gpt-5.4} incorporates few-shot explanations and the best performing prompt for gratitude with \courier{llama3.3} incorporates a persona. Yet, in most cases, additive techniques offer little or no benefit beyond the baseline prompt. Overall, the difference between the bottom baseline prompt and prompts incorporating additive techniques is never statistically significant, though at a maximum, chain-of-thought prompting increased the F1 score for negative core beliefs from 0.51 to 0.60 with \courier{llama3.3}.

\subsection*{Summary of Results}

\begin{figure}[h!]
    \centering
    \caption{\textit{Summary of Prompting Technique Effectiveness, As Estimated in the Development Dataset}}
    \includegraphics[width=0.7\textwidth]{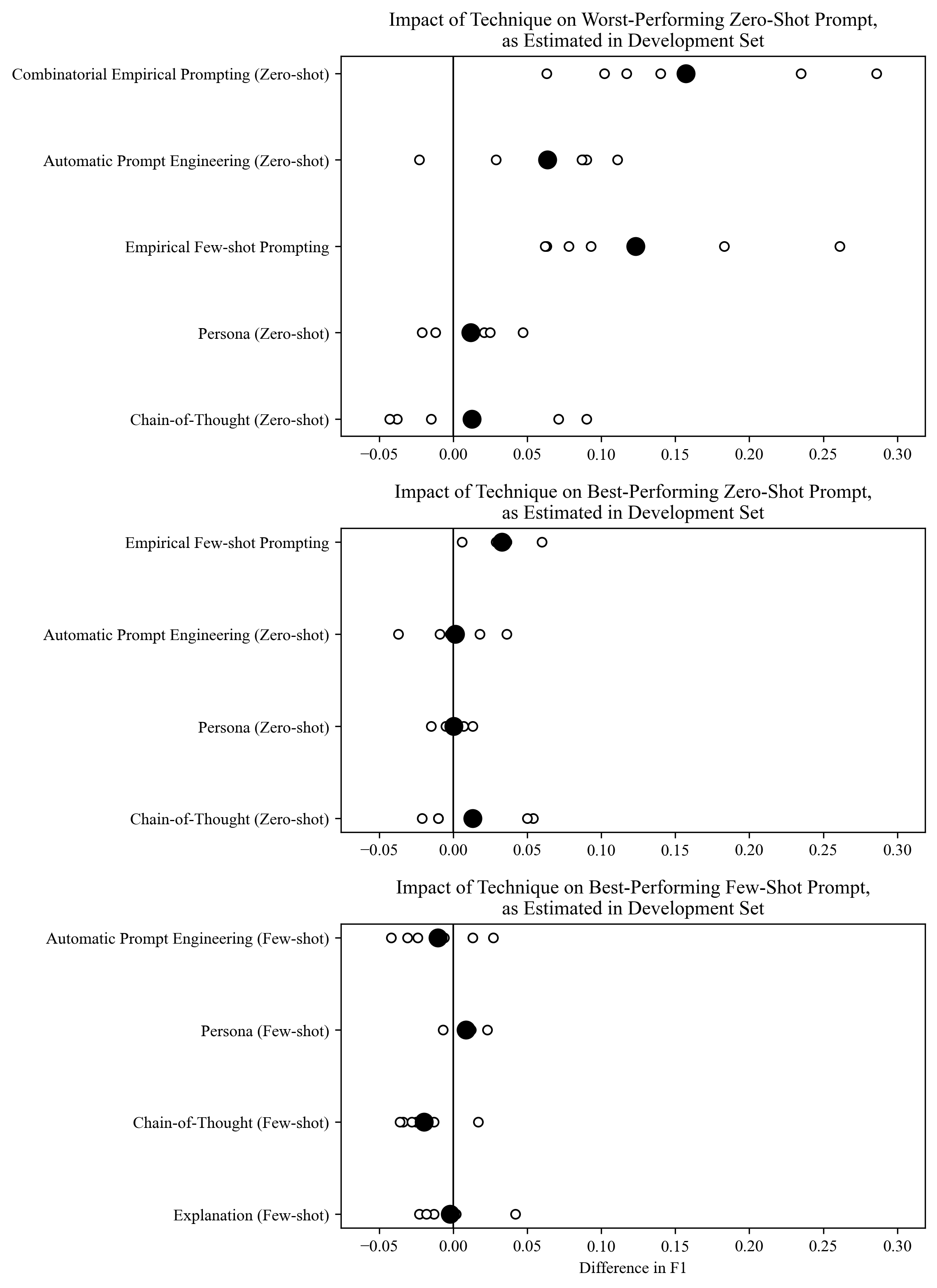}
    \label{fig:c}
\end{figure}

We summarize our findings and related recommendations in Figure \ref{fig:c}, with vertical panels ordered by the stage of the prompt engineering process. The empty dots display the difference in F1 scores due to a given technique for each construct-model combination, while the filled dots display the average difference in performance across construct-model combinations. The top panel addresses the following question: which technique most improves an ineffective baseline prompt? The figure shows that combinatorial empirical prompting makes the greatest difference, generating an average increase of 0.17 in F1 score, estimated in the development dataset. The middle panel then addresses a second-order question: after identifying an effective baseline prompt with combinatorial empirical prompting, what technique is most effective next? Here, the answer is empirical few-shot prompting, which increases F1 by 0.03 on average. 

Finally, the bottom panel addresses: given an empirically selected prompt and few-shot examples, to what extent can performance be improved with automatic prompt engineering or additive prompts? Results in this panel are mixed at best. On average, automatic prompt engineering, chain-of-thought prompting, and explanations slightly reduce performance, though by negligible amounts. Personas, on the other hand, increase F1 on average, but only by 0.01. Overall, mean differences are centered near zero, but there are individual cases where incorporating an additive technique results in meaningful improvement. For example, incorporating explanations into the negative core beliefs prompt with \courier{gpt-5.4} resulted in a 0.04 increase in F1. 

\subsection*{Validation in the Testing Dataset}

Based on performance in the development dataset, we select the top-performing prompt–model combination for validation in the hold-out testing dataset. Our final classifiers use \courier{gpt-5.4} because the model outperformed \courier{llama3.3}. For gratitude, the highest performing prompt is the empirically selected few-shot prompt with no additive techniques, achieving an estimated specificity of 0.95, recall of 0.89, and precision of 0.95. For meaning making, the highest-performing prompt combined the top-performing combinatorial prompt with automatic prompt engineering and empirically selected few-shot examples. This resulted in a specificity of 0.92, recall of 0.94, and a precision of 0.77. Finally, for negative core beliefs, the highest-performing combination is the empirically selected few-shot prompt with explanations, achieving an estimated specificity of 0.95, recall of 0.84, and a precision of 0.85. 

In addition to assessing alignment with human classifications, we also examine a random sample of 15 disagreements and 15 agreements for each construct to identify themes and patterns. Within the gratitude dataset, we find that human coders are more generous in their classifications of gratitude, often classifying texts expressing general positive sentiment as gratitude. For example, two of our random disagreements included posts that simply wished another user happy birthday; while the LLM did not classify these as gratitude, the human did. For meaning making, we find that disagreements often occur when the described positive impact is relatively limited compared to the described negative impact. For example, when texts were largely focused on negative impacts but included comments like “[showed] me that I need to make my own decisions” or “shown me that I can get through anything”, human coders did not classify these as meaning making, but the model did. Though such comments are positive benefits, the remainder of the text does not suggest that these benefits are part of a larger “restoration of meaning” (\cite{park_making_2010}) and, overall, human coders appeared to require a more substantial reinterpretation of the negative experience. 

For negative core beliefs, patterns in disagreement are less clear-cut. In the sampled cases, however, human coders appear more willing to infer generalized beliefs from behavioral evidence (e.g., “don’t like showing my vulnerability”, “do not share anything”) whereas the model appears more likely to classify negative self-evaluations as negative core beliefs, even when those statements do not fit the common template for negative core beliefs (i.e., “I am…”, “People are…”) and are not clearly generalized. For example, human coders did not classify texts including statements like “whatever I do has to be perfect or it's not acceptable,” and “[I] feel like I would get judged [for] how I react to things” as negative core beliefs but the model did. Overall, these exploratory observations suggest that disagreements often reflect 1) differing thresholds for inference and 2) differences in the weight given to different kinds of textual evidence in support of those inferences, both elements that can be influenced by prompting. 

\begin{table}
\caption{\textit{Final Performance Metrics}}
\begin{center}
\begin{tabular}{l c c c c c}
    \hline
     Concept & Accuracy & Specificity & Recall & Precision & F1\\
        \hline
        Gratitude & 0.92 & 0.95 & 0.89 & 0.95 & 0.92\\
         & [0.88, 0.95] & [0.91, 0.98] & [0.84, 0.94] & [0.91, 0.98] & [0.88, 0.95]\\
        Meaning Making & 0.92 & 0.92 & 0.94 & 0.77 & 0.85\\
         & [0.88, 0.96] & [0.87, 0.96] & [0.86, 1.00] & [0.65, 0.88] & [0.76, 0.92]\\
        Negative Core Beliefs & 0.92 & 0.95 & 0.84 & 0.85 & 0.85\\
         & [0.89, 0.96] & [0.92, 0.98] & [0.72, 0.93] & [0.74, 0.95] & [0.76, 0.92]\\
        \hline
\end{tabular}
\end{center}
\textit{Note.} Confidence intervals are in brackets and calculated via bootstrapping.
\label{table:9}
\end{table}

\section*{Discussion and Recommendations}

If validity is understood as the “approximate truth of an inference” (\cite{shadish_experimental_2001}, p. 34), then the validity of LLM-based classifiers concerns the approximate truth of their resulting classifications (e.g., does this text really indicate that the client holds a negative core belief?). Yet, when psychological constructs are latent, such inferences cannot be proven (e.g., we cannot know for certain what beliefs a client truly holds, no matter what they write). Instead, a fundamental source of validity evidence for LLM-based classifications is their alignment with human experts, under the assumption that expert classifications have validity of their own. 

Across our experiments, our findings reinforce what prior literature has emphasized; prompting matters for increasing alignment with expert codes (\cite{sahoo_systematic_2024}; \cite{weber_evaluation_2023}). Our findings also point to the importance of both theory-driven prompt generation and empirical experimentation. First, our results suggest that theory-driven prompt components like construct definitions, inclusion/exclusion criteria, and examples are more influential than LLM-specific prompting techniques like automatic prompt engineering, personas, chain-of-thought, or explanations. However, which specific combinations of components and examples are most effective is not always obvious a priori. Combinatorial empirical prompting addresses this challenge by drawing on the human codebook for content and using empirical experimentation for prompt and example selection. 

Our results also suggest small and inconsistent benefits of automatic prompt engineering, personas, chain-of-thought, and explanations. These techniques often improve upon an ineffective prompt, though not to the same extent as combinatorial empirical prompting or few-shot example selection. Nonetheless, there are cases where the highest-performing prompt incorporates these techniques: our final prompt for meaning making incorporates automatic prompt engineering and our final prompt for negative core beliefs incorporates explanations. 

Given the above results, when using LLMs to identify constructs in text, we recommend that researchers consider the following process for assessing and improving the alignment between LLM and human classifications:

\begin{enumerate}
    \item Generate a comprehensive construct codebook, where, as much as possible, rules and guidance about reasonable inferences and appropriate evidence are made explicit. 
    \item Generate a diverse set of baseline prompt variants by combining alternative versions of (a) the construct definition, (b) task instructions, and (c) inclusion and exclusion criteria, including variants with no inclusion or exclusion criteria and variants with all available criteria. Evaluate prompts from this pool in the training dataset and select the highest-performing combination.
    \item If resources permit, siphon off a portion of the training data for few-shot examples. From this example pool, generate a diverse set of few-shot combinations varying the number of examples, ratio of positive to negative examples, and the examples themselves. Append each combination to the best-performing baseline prompt and evaluate these few-shot variants in the training dataset, excluding the example pool. Identify the top-performing combination and compare performance between the zero-shot and few-shot approaches, again excluding the example pool from this evaluation.
    \item Again, if resources permit, experiment with appending additive techniques to the current best-performing prompt, including the additive techniques discussed in this article (persona, chain-of-thought, and/or explanations), other techniques not discussed, or new alternative techniques as they arise.
    \item Conduct a final evaluation in a hold-out testing dataset, calculating accuracy, specificity, precision, recall, and F1 score, alongside associated confidence intervals to provide an unbiased estimate of human-machine alignment.
\end{enumerate}

This procedure directs effort toward the aspects of prompt engineering that our findings suggest most strongly influence alignment with human classifications—prompt wording, structure, and example selection—while emphasizing empirical validation of the resulting classifications.

\subsection*{Limitations}

While we believe the above recommendations provide a reasonable and empirically grounded process, our experiments, and therefore our conclusions, come with important limitations.  First and foremost, our development datasets are not large enough to reliably identify small prompting effects. For example, due to class imbalance, there are just 65 examples of negative core beliefs in our development dataset (290 labeled texts total). Further, because participants often generated multiple texts (approximately two on average), observations are not independent, an issue we address through hierarchical bootstrapping. Together, these features – imbalanced classes and a nested data structure – result in moderately sized standard errors that only allow us to identify relatively large prompting effects with statistical significance. This is one reason we emphasize comparative patterns across techniques and case studies rather than relying solely on null hypothesis significance testing. Likewise, we do not conclude that the tested additive techniques are ineffective, only that, when resources are limited, it is likely more efficient to prioritize combinatorial baseline prompting and few-shot example selection. Further, we hope that the inclusion of three case studies, and the assessment of generalizability across those case studies, helps temper limitations associated with moderate sample sizes. 

Second, our findings come with generalizability challenges; what works in one scenario may not work in another (\cite{battle_unreasonable_2024}). Again, we hope that our inclusion of multiple constructs helps address this limitation. Relatedly, model capabilities are certain to change over time, and new prompting techniques are likely to emerge, potentially limiting the generalizability of our specific findings across model generations. However, even as LLMs evolve, we expect the broader process of systematic prompt generation, empirical evaluation, and validation to remain relevant. Third, we should also acknowledge that prompt engineering, as we recommend it, is a time-consuming process that should only be undertaken when resources allow and when maximizing classification performance and validity is important. If LLM-based classification will be used to support or guide clinical decision-making, this investment will likely be worthwhile.

Finally, we also hope that future research addresses key questions raised here, including: 1) whether and when overfitting during prompt engineering is enough of a concern to warrant a separation between the training and development data (as our results suggest uncertainty due to sampling variation is the greater concern); and 2) minimum and recommended sample sizes for training (identifying the best of a given number of prompts) and hold-out validation (e.g., producing metrics with confidence intervals that are sufficiently narrow for decision-making). 

\section*{Conclusion}

Across three constructs and multiple datasets, we find that alignment between LLM and human classifications depends far more on baseline prompt composition and example selection than on widely discussed additive prompting techniques. Combinatorial empirical prompting—systematically generating and evaluating alternative combinations of construct definitions, task instructions, and coding guidance—substantially improved alignment with human classifications, as did the empirical selection of examples. More broadly, we demonstrate and test a process of prompt engineering for psychological construct identification that systematically combines theoretical specification, experimentation, and validation. As LLMs become increasingly integrated into psychological and behavioral research, these practices will be essential for producing classifications that can be trusted for use in decision making.

\section*{Open Practices}
All analysis code is publicly available at \url{https://github.com/KylieLAnglin/pe4ci}. Due to the sensitive nature of the content and to protect participant privacy, two of the datasets used in this study—those related to meaning making and negative core beliefs—are available at reasonable request. However, the dataset used for the gratitude construct (\cite{demszky_goemotions_2020}) is available at \url{https://github.com/google-research/google-research/tree/master/goemotions} and can be used to replicate and illustrate the analysis procedures.

\section*{Declarations}

\subsection*{Funding} 
Some of the data collected within this study was funded with support from the University of Connecticut Dean’s Office Summer Research Awards.

\subsection*{Conflicts of Interest} 
The authors declare that they have no competing interests.

\subsection*{Ethics Approval}
The data used for analyses were collected from studies approved by the University of Connecticut IRB (IRB protocol \#H23-0244 and \#X18-057).

\subsection*{Consent to Participate}
Individuals in these studies provided informed consent prior to participation.

\subsection*{Consent for Publication}
Individuals in these studies provided informed consent prior to participation.

\newpage
\printbibliography

\newpage
\section*{Appendix}
\subsection*{Appendix A}

\renewcommand{\tablename}{Table}

\renewcommand{\thetable}{A1}

\begin{table}[H]
    \caption{\textit{Comparison of OpenAI Models and Settings}}
    \begin{center}
    \begin{tabular}{l c c c c c}
        \hline
        Model & Effort & Temperature & F1 & $r^2$ & \shortstack{Average Output and Reasoning\\Tokens Per Text}\\
            \hline
            gpt-5.4 & none & & 0.89 & 0.97 & 7\\
            gpt-5.4 & low & & 0.87 & 0.88 & 196\\
            gpt-5.4 & medium & & 0.86 & 0.82 & 361\\
            gpt-4.1-mini-2025-04-14 & & 1 & 0.83 & 1.00 & 3\\
            gpt-5-mini & minimal & & 0.83 & 0.74 & 21\\
            gpt-4.1-mini-2025-04-14 & & 0 & 0.82 & 1.00 & 3\\
            gpt-4o-mini-2024-07-18 & & 1 & 0.82 & 0.97 & 3\\
            gpt-5-mini & medium & & 0.82 & 0.89 & 534\\
            gpt-4.1-mini-2025-04-14 & & 0.3 & 0.82 & 0.97 & 3\\
            gpt-5-mini & low & & 0.81 & 0.65 & 230\\
            gpt-4o-mini-2024-07-18 & & 0 & 0.80 & 0.93 & 3\\
            gpt-4o-mini-2024-07-18 & & 0.3 & 0.80 & 0.96 & 3\\
            gpt-4.1-nano-2025-04-14 & & 0 & 0.80 & 0.97 & 58\\
            gpt-5-nano & minimal & & 0.78 & 0.78 & 23\\
            gpt-4.1-nano-2025-04-14 & & 1 & 0.78 & 0.85 & 57\\
            gpt-5-nano & low & & 0.77 & 0.41 & 235\\
            gpt-4.1-nano-2025-04-14 & & 0.3 & 0.77 & 0.94 & 58\\
            gpt-5-nano & medium & & 0.76 & 0.86 & 1259\\     
            \hline
    \end{tabular}
    \end{center}
    \label{table:app-a}
\end{table}

\begin{landscape}
\subsection*{Appendix B}
\subsubsection*{Exploratory Analysis of Estimated Impact of Prompt Components on Precision, Recall, and F1 in Training Data}

\renewcommand{\thetable}{B1}

\begin{longtable}{l c c c c c c p{11cm}}
    \caption{\textit{Relative Impact of Gratitude Prompt Components within OpenAI Model, as Estimated in the Training Data}\label{table:app-b1}}\\
    \hline
    \shortstack{Component\\Type} & \shortstack{Component\\ID} & \shortstack{Difference\\in Recall} & \shortstack{Difference in \\Precision} & \shortstack{Difference in \\F1} & \shortstack{Included\\in Top\\Prompt} & \shortstack{Included\\in Bottom\\Prompt} & Text\\ 
    \hline
            context & 0 & -0.01 & 0 & 0 & 0 & 1 & The following text is from a social media post.\\
            context & 1 & 0 & 0 & 0 & 0 & 0 & The following text is from Reddit.\\
            context & 2 & 0.01 & 0 & 0.01 & 1 & 0 & \\
            \hline
            task & 0 & 0.01 & 0 & 0.01 & 0 & 0 & Your task is to determine whether the text indicates gratitude from the author.\\
            task & 1 & 0 & 0 & 0 & 0 & 0 & Your task is to classify the text as expressing gratitude or not.\\
            task & 2 & 0 & 0 & 0 & 1 & 0 & Your task is to decide whether the text suggests that the author is grateful.\\
            task & 3 & -0.01 & 0 & -0.01 & 0 & 1 & Your task is to determine whether this reddit post suggests the author is grateful.\\
            \hline
            definition & 0 & 0 & -0.01 & 0 & 0 & 0 & Gratitude is defined as feeling of appreciation for others or circumstances.\\
            definition & 1 & 0 & 0 & 0 & 1 & 0 & Gratitude is the feeling of being thankful.\\
            definition & 2 & 0 & 0 & 0 & 0 & 1 & Gratitude is a sense of thankfulness and happiness in response to receiving a gift, either a tangible benefit (e.g., a present, favor) given by someone or a fortunate happenstance (e.g., a beautiful day).\\
            definition & 3 & 0 & 0 & 0 & 0 & 0 & Gratitude is a feeling of thankfulness and appreciation.\\
            \hline
            guidance & 1 & -0.01 & 0 & 0 & 1 & 1 & Some authors explicitly state that they are thankful. Others only imply gratitude.\\
            guidance & 2 & -0.01 & 0 & 0 & 0 & 1 & The text may indicate other emotions in addition to gratitude. For example, they may be joyful and thankful. In this case, you should still classify as gratitude.\\
            guidance & 3 & 0 & 0 & 0 & 0 & 1 & If you think the person is or was grateful regarding the topic, classify as gratitude.\\
            guidance & 4 & 0 & 0 & 0 & 1 & 1 & Some words that indicate gratitude include grateful, thank you, sharing, and blessed.\\
            guidance & 5 & 0.01 & 0 & 0.01 & 1 & 0 & Pay particular attention to positive words.\\
            guidance & 6 & -0.01 & 0.01 & 0 & 0 & 1 & Some positively toned texts will just be happy but not grateful.\\
            guidance & 7 & -0.04 & 0 & -0.03 & 0 & 1 & Watch out for sarcasm. Sarcastic thank yous should not be classified as gratitude.\\
            guidance & 8 & -0.02 & 0.01 & -0.01 & 0 & 1 & If you doubt that the person is truly grateful, do not classify as gratitude.\\
            \hline
\end{longtable}

\newpage
\renewcommand{\thetable}{B2}
\begin{longtable}{l c c c c c c p{11cm}}
    \caption{\textit{Relative Impact of Meaning Making Prompt Components within OpenAI Model, as Estimated in the Training Data}\label{table:app-b2}}\\
    \hline
    \shortstack{Component\\Type} & \shortstack{Component\\ID} & \shortstack{Difference\\in Recall} & \shortstack{Difference in \\Precision} & \shortstack{Difference in \\F1} & \shortstack{Included\\in Top\\Prompt} & \shortstack{Included\\in Bottom\\Prompt} & Text\\
    \hline
    \endfirsthead
    \hline
    \shortstack{Component\\Type} & \shortstack{Component\\ID} & \shortstack{Difference\\in Recall} & \shortstack{Difference in \\Precision} & \shortstack{Difference in \\F1} & \shortstack{Included\\in Top\\Prompt} & \shortstack{Included\\in Bottom\\Prompt} & Text\\
    \hline
    \endhead
        context & 0 & 0 & 0.02 & 0.01 & 1 & 0 & As part of a therapeutic program, clients have been asked to write in responses to questions recent experiences and their childhood, including positive and negative experiences.
        You are about to read a writing sample in response to one of those questions.\\
        context & 1 & 0.01 & 0.01 & 0.01 & 0 & 1 & The following text is a short personal reflection written by an individual in response to open-ended survey questions about the impact of past positive or negative experiences.\\
        context & 2 & 0 & -0.01 & 0 & 0 & 0 & In the following text, an individual responds to questions about their past experiences.\\       
        context & 3 & -0.02 & -0.02 & -0.02 & 0 & 0 & \\
        \hline
        task & 0 & 0.01 & 0 & 0.01 & 1 & 0 & Your task is to determine whether the text indicates that the author has engaged in meaning making in response to a negative life event.\\
        task & 1 & 0.02 & -0.02 & -0.01 & 0 & 0 & Your task is to determine whether the text contains evidence of meaning making.\\
        task & 2 & -0.03 & 0.01 & 0 & 0 & 0 & Your task is to decide whether there is sufficient evidence in the text to confidently conclude that the author has engaged in meaning making in response to a negative experience.\\
        task & 3 & -0.01 & 0.01 & 0 & 0 & 1 & Your task is to determine whether the text reflects meaning making -- identifying a positive change due to a negative experience.\\
        \hline
        definition & 0 & 0.02 & 0 & 0.01 & 0 & 0 & When someone engages in meaning making, they find an important benefit of a negative event or experience.\\
        definition & 1 & 0.02 & -0.03 & -0.01 & 0 & 0 & Meaning making refers to when a person reports some positive internal change or outcome resulting from a negative external event, trauma, or stressor in their life.\\
        definition & 2 & 0 & -0.01 & 0 & 0 & 1 & Positive meaning-making refers to the psychological process of finding, creating, or interpreting meaning in response to negative life events in a way that enhances well-being.\\
        definition & 3 & -0.03 & -0.02 & -0.03 & 0 & 0 & The psychological process of meaning making occurs when someone assigns meaning to a highly stressful negative life experience, bringing it into a more positive narrative of their life.\\
        definition & 4 & -0.01 & 0.04 & 0.02 & 1 & 0 & Meaning making occurs when someone describes how they have grown in response to a negative event, incorporating the negative event into a more positive narrative of their life.\\          
        \hline
        guidance & 1 & 0 & 0.01 & 0 & 1 & 0 & Meaning making is often understood as a type of growth.\\
        guidance & 2 & 0.01 & 0.01 & 0.01 & 0 & 0 & Typically, meaning-making is reflected in a person describing a change in personal attributes (e.g., becoming a stronger or more empathic person), self-view (e.g., having more belief in oneself), relationships (e.g., “it brought us closer”), or the acquisition of valuable lessons (e.g., “life is short and should be lived fully, day to day”).\\
        guidance & 3 & 0.01 & -0.03 & -0.01 & 0 & 0 & In expressive-writing interventions, meaning-making is conceptualized as evidence of cognitive processing or reappraisal.\\
        guidance & 4 & 0 & 0.01 & 0 & 1 & 0 & Some examples of phrases associated with meaning-making include: “led me to be more accepting of things,” “taught me how to adjust to things I cannot change,” “taught me to be patient,” “brought my family closer together,” “helped me become a stronger person,” “helped me grow emotionally and spiritually,” “made me more compassionate toward those in similar situations,” “taught me that everyone has a right to be valued,” “led me to place less emphasis on material things,” and “led me to change my priorities in life.”\\
        guidance & 5 & 0.01 & 0.01 & 0.01 & 0 & 0 & What you are looking for as evidence of positive meaning-making is a negative–positive event link (a negative past event leading to something positive derived from that event). The “positive” may be a beneficial change in the individual, a valued lesson or insight, or a behavior they view as positive that they engage in specifically because they want to be different from what happened to them.\\
        guidance & 6 & -0.03 & 0 & -0.01 & 0 & 1 & To qualify as meaning-making, the growth must result from a negative event.\\
        guidance & 7 & 0 & 0.03 & 0.02 & 1 & 0 & Sometimes people report positive outcomes from treatment following a negative event (e.g., benefits of therapy). These do not qualify as meaning-making because the treatment itself is not a negative event.\\
        guidance & 8 & -0.01 & 0.03 & 0.01 & 0 & 0 & Meaning making should not be coded simply because the person is doing better now or because the event serves as a contrast to a better present (e.g., “That was a bad time in my life; things are much better now. I know I never want to go back to that time”). The response must explicitly connect the growth or learning to the event (i.e., a change in attributes, mindset, or lessons resulting from the event).\\
        guidance & 9 & 0 & 0.03 & 0.02 & 1 & 0 & Meaning making requires that the negative event is the cause of the positive outcome.\\
        guidance & 10 & -0.01 & 0 & -0.01 & 0 & 0 & Some people phrase things in a “growth” way, but it is unclear if this is positive (e.g., “It taught me that nobody can be trusted so only rely on yourself”). If the growth does not appear to be genuine or positive, it does not qualify as meaning making.\\
        guidance & 11 & -0.04 & 0.03 & 0 & 0 & 0 & Changes such as “I’m less naïve,” “I now know that people can be [negative trait],” “I don’t rely on others,” or “I now know the world can be unfair” do not qualify as positive benefits or growth and should not be used to classify a response as meaning-making.\\
        guidance & 12 & 0 & 0.01 & 0.01 & 0 & 0 & It is not necessary for all outcomes to be positive to qualify as meaning-making. The author may describe both negative and positive impacts. As long as the positive impacts are genuinely positive, valued by the author, and not outweighed by the described negative impacts, the response qualifies as meaning making.\\
        \hline  
\end{longtable}

\newpage
\renewcommand{\thetable}{B3}
\begin{longtable}{l c c c c c c p{11cm}}
    \caption{\textit{Relative Impact of Negative Core Beliefs Prompt Components within OpenAI Model, as Estimated in the Training Data}\label{table:app-b3}}\\
    \hline
    \shortstack{Component\\Type} & \shortstack{Component\\ID} & \shortstack{Difference\\in Recall} & \shortstack{Difference in \\Precision} & \shortstack{Difference in \\F1} & \shortstack{Included\\in Top\\Prompt} & \shortstack{Included\\in Bottom\\Prompt} & Text\\
    \hline
    \endfirsthead
    \hline
    \shortstack{Component\\Type} & \shortstack{Component\\ID} & \shortstack{Difference\\in Recall} & \shortstack{Difference in \\Precision} & \shortstack{Difference in \\F1} & \shortstack{Included\\in Top\\Prompt} & \shortstack{Included\\in Bottom\\Prompt} & Text\\
    \hline
    \endhead
        context & 0 & 0.1 & -0.09 & -0.01 & 0 & 0 & As part of a therapeutic program, clients have been asked to write in responses to questions about their identities and their childhood, including positive and negative experiences.
        You are about to read a writing sample in response to one of those questions.\\
        context & 1 & -0.02 & 0.03 & 0 & 0 & 1 & The following text is a short personal reflection written by an individual in response to open-ended survey questions about the impact of past positive or negative experiences.\\
        context & 2 & -0.03 & 0.03 & 0.01 & 1 & 0 & In the following text, an individual responds to questions about their past experiences.\\
        context & 3 & -0.02 & 0.02 & 0 & 0 & 0 & \\
        \hline
        task & 0 & 0.06 & -0.04 & 0.02 & 0 & 0 & Your task is to determine whether the text either strongly implies or explicitly states a negative core belief about the self or others currently held by the author.\\
        task & 1 & -0.12 & 0.1 & -0.02 & 0 & 0 & Your task is to determine whether the text contains strong evidence that the author currently holds a negative core belief.\\
        task & 2 & 0.01 & -0.01 & -0.01 & 0 & 1 & Your task is to decide whether there is sufficient evidence in the text to confidently conclude that the author currently holds a negative core belief about themselves or others.\\
        task & 3 & 0.02 & -0.03 & 0 & 1 & 0 & Your task is to determine whether the text reflects one or more, specific, negative core beliefs.\\
        \hline
        definition & 0 & 0.02 & -0.04 & -0.01 & 0 & 0 & Negative core beliefs are broad, generalized, or exaggerated negative beliefs that people hold about themselves or others.\\
        definition & 1 & 0.07 & -0.04 & 0.02 & 0 & 0 & Negative core beliefs are broad, persistent negative beliefs a person holds about themselves, others, the world, or the future.\\
        definition & 2 & -0.09 & 0.06 & -0.02 & 0 & 1 & A negative core belief is a strongly held and broad negative belief. \\
        definition & 3 & 0.02 & 0 & 0.01 & 1 & 0 & Negative core beliefs are beliefs about the self, others, the world, or future that are deeply held, persistent, pervasive, and negative.\\
        \hline
        guidance & 1 & 0.08 & -0.06 & 0.01 & 1 & 0 & The negative core belief may not be explicitly stated, but to code “yes,” there should be enough evidence to form a strong hypothesis of what the specific negative core belief(s) might be (e.g., “People are \underline{\hspace{1cm}}.” “I am \underline{\hspace{1cm}}.”). One exception to this rule is that, if the writer states that they hold negative beliefs about themselves or others, code as a negative core belief even though we may not know what those beliefs are.\\
        guidance & 2 & -0.03 & 0.03 & 0.01 & 0 & 0 & To determine if there is sufficient evidence of a negative core belief, you might try to complete one of these sentences: “I am \underline{\hspace{1cm}},” “People are \underline{\hspace{1cm}},” or “People will \underline{\hspace{1cm}},” where people may instead be a group of people like men.\\
        guidance & 3 & -0.06 & 0.03 & -0.01 & 0 & 0 & Look for global self-statements (e.g., “I am …”) or generalized statements about groups (e.g., people, men).\\
        guidance & 4 & 0.02 & 0 & 0.01 & 0 & 1 & Negative core beliefs often result from generalizing far beyond an original offender.\\
        guidance & 5 & 0.04 & -0.02 & 0.01 & 0 & 0 & It is possible for the text to be extremely negative without providing strong evidence of a negative core belief about the self or others.\\
        guidance & 6 & -0.03 & 0.05 & 0.02 & 1 & 0 & Observations about the author’s behavior, even if that behavior reflects trauma, do not by themselves qualify as a negative core belief. They can, however, provide supporting evidence when combined with other statements.\\
        guidance & 7 & -0.02 & 0.04 & 0.01 & 1 & 1 & If the author talks about past negative core beliefs, that should not be coded as “yes” unless there is evidence that the belief is still held.\\
        guidance & 8 & -0.13 & 0.11 & -0.01 & 1 & 1 & Negative statements that apply to either specific individuals or to “some people” do not imply negative core beliefs. Negative core beliefs must be generalized to many people, or a large group of people.\\
        guidance & 9 & -0.02 & 0 & -0.01 & 0 & 0 & Qualifiers like “some people,” “certain people,” “sometimes,” or “at times” temper the statement so that it is less generalized. Qualifiers that imply more than half of people (or half of a group of people) or that imply more than half of the time are considered generalized.\\
        guidance & 10 & -0.12 & 0.1 & 0 & 1 & 1 & Conditional statements should also be treated as qualifiers, tempering the belief so that the statement alone is not generalized enough to qualify as a negative core belief (e.g., “I think I am weak if I show emotions”; “People who \underline{\hspace{1cm}} are…”).\\
        guidance & 11 & -0.09 & 0.07 & 0 & 0 & 1 & Descriptions of behavior like “I have a hard time/I struggle with trusting people” should be treated as a qualifier. Whether there is enough evidence of a negative core belief depends on the rest of the text.\\
        guidance & 12 & 0.08 & -0.04 & 0.02 & 0 & 0 & People don’t often explicitly state the generalized belief. Instead, they express it as an ongoing fear, thought, feeling, or worry. \\
        guidance & 13 & 0.05 & -0.02 & 0.02 & 0 & 0 & When people write “I’m afraid that” or “I’m worried that,” translate this to “I believe that” so that if the rest of the sentence is generalized, it may indicate a negative core belief.\\
        guidance & 14 & 0.11 & -0.1 & 0 & 0 & 0 & Expressing fear of showing emotions or being vulnerable is evidence in favor of a negative core belief. Expressing not knowing how to show emotion or just disliking it is not.\\
        guidance & 15 & 0.06 & -0.02 & 0.03 & 1 & 0 & Expressing distrust of groups of people, or people in general, is evidence of a negative core belief.\\
        \hline
\end{longtable}

\end{landscape}

\subsection*{Appendix C}
\subsubsection*{Evolution of Performance with Automatic Prompt Engineering}
\renewcommand{\figurename}{Appendix}
\renewcommand{\thefigure}{C}

\begin{figure}[h!]
    \centering
    \caption{\textit{Performance Evolution with Automatic Prompt Engineering, as Assessed in the Training Dataset, with Highest F1 Indicated with Bold}}
    \includegraphics[width=\textwidth]{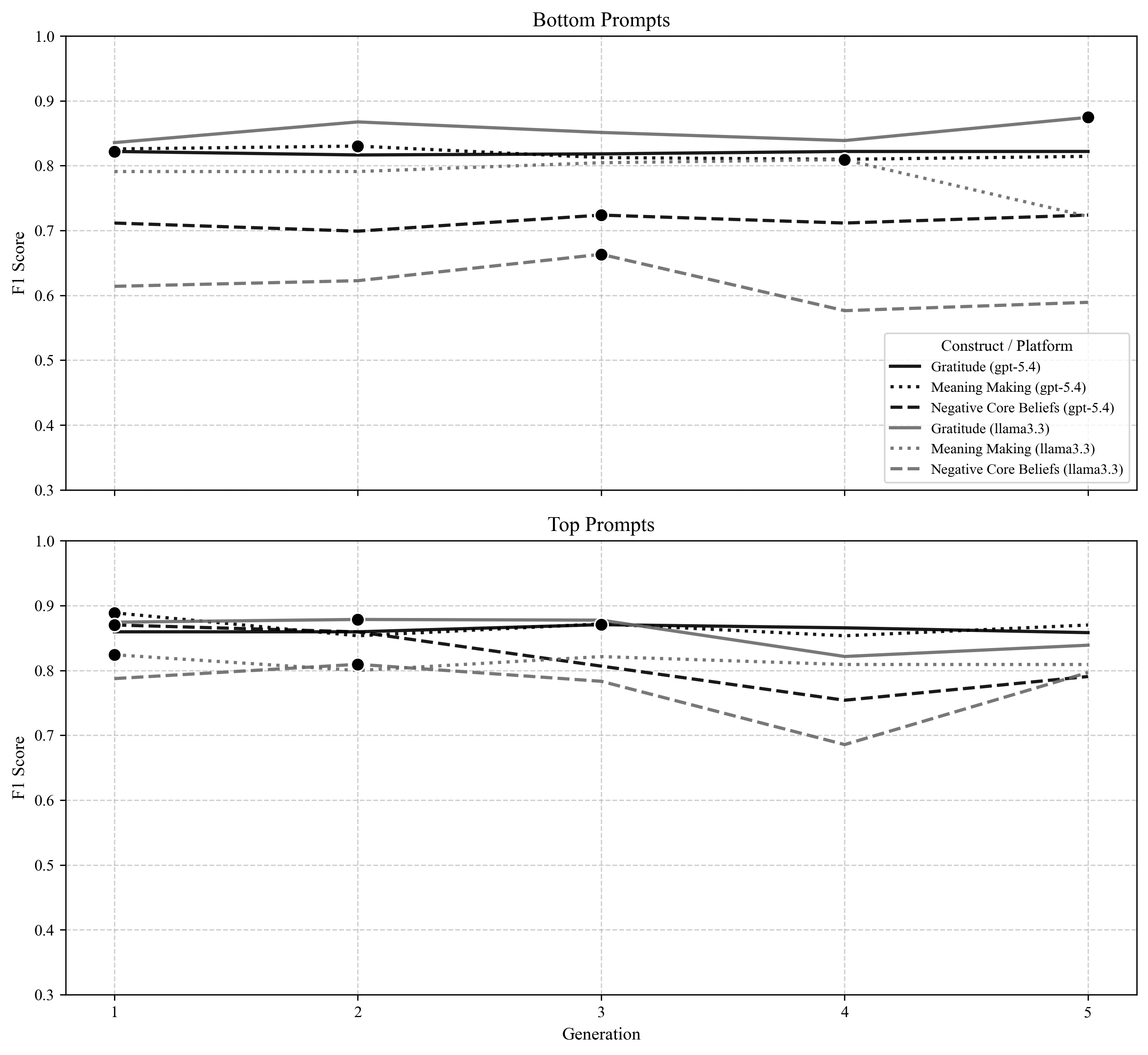}
    \label{fig:app-c}
    \textit{Note.} Black markers indicate the highest F1 for a given construct-platform combination.
\end{figure}

\end{document}